\documentclass[runningheads]{llncs}
\usepackage{graphicx}
\usepackage{amsfonts}
\usepackage{amsmath}
\usepackage{multirow}
\usepackage{subfigure}

\usepackage{amssymb}
\usepackage{makecell}
\usepackage{framed,color}
\usepackage{booktabs}
\usepackage{marvosym}
\usepackage{stfloats}
\usepackage{bm}
\begin{document}
\title{MM-UNet: A Mixed MLP Architecture for Improved Ophthalmic Image Segmentation}

\titlerunning{MM-UNet for Improved Ophthalmic Image Segmentation}
%
%
%
%

\author{Zunjie Xiao\inst{1} \and
Xiaoqing Zhang\inst{1} \and
Risa Higashita \inst{1,2}\textsuperscript{\Letter} \and
Jiang Liu\inst{1,3}\textsuperscript{\Letter}}
\authorrunning{Z. Xiao et al.}
%
\institute{Research Institute of Trustworthy Autonomous Systems and Department of Computer Science and En-gineering, Southern University of Science and Technology, Shenzhen, 518055, China
\and 
TOMEY Corporation, Nagoya, 4510051, Japan.
\and School of Computer Science, University of Nottingham Ningbo China, Ningbo 315100, China
\\
\email{liuj@sustech.edu.cn}
}     
\maketitle              
\begin{abstract}

Ophthalmic image segmentation serves as a critical foundation for ocular disease diagnosis. Although fully convolutional neural networks (CNNs) are commonly employed for segmentation, they are constrained by inductive biases and face challenges in establishing long-range dependencies. Transformer-based models address these limitations but introduce substantial computational overhead. Recently, a simple yet efficient Multilayer Perceptron (MLP) architecture was proposed for image classification, achieving competitive performance relative to advanced transformers. However, its effectiveness for ophthalmic image segmentation remains unexplored.
In this paper, we introduce MM-UNet, an efficient Mixed MLP model tailored for ophthalmic image segmentation. Within MM-UNet, we propose a multi-scale MLP (MMLP) module that facilitates the interaction of features at various depths through a grouping strategy, enabling simultaneous capture of global and local information. We conducted extensive experiments on both a private anterior segment optical coherence tomography (AS-OCT) image dataset and a public fundus image dataset. The results demonstrated the superiority of our MM-UNet model in comparison to state-of-the-art deep segmentation networks.
\end{abstract}
\section{Introduction}

In clinical diagnosis, the segmentation of ophthalmic images is a critical step \cite{zhang2022attention,zhang2024regional}. Many automatic measurements of ophthalmic clinical parameters rely on accurate segmentation results \cite{wang2020artificial}. Early research primarily utilized traditional image processing algorithms for segmentation, such as Canny edge detection \cite{cheng2010closed,li2010computer}, often accompanied by complex image preprocessing. With the success of deep learning across various fields, Convolutional Neural Networks (CNNs) have demonstrated substantial progress in many ophthalmic segmentation tasks \cite{cao2020efficient,sevastopolsky2017optic,saravanan2022deep}, ranging from lens segmentation \cite{cao2020efficient} in the ophthalmic anterior segment to disc segmentation \cite{saravanan2022deep} in the posterior segment. However, CNN-based models generally exhibit limitations in establishing long-range dependencies, as convolution operations primarily capture local information. Consequently, CNN-based approaches often perform inadequately on target structures with diverse shapes and textures.


The introduction of transformer architectures \cite{vaswani2017attention} provides a practical solution to overcome these limitations, with examples including ViTs \cite{dosovitskiy2020image}, TransUNet \cite{chen2021transunet}, and UTNet \cite{gao2021utnet}. These transformer-based methods have also achieved significant performance in ophthalmic image segmentation \cite{zhang2021multi}. In recent years, it has been commonly believed that the self-attention mechanism is necessary for establishing long-range dependencies in deep networks. However, the recently proposed pure multi-layer perceptron (MLP) models, such as the MLP-Mixer \cite{tolstikhin2021mlp}, demonstrate that a simple stack of MLPs also holds great potential. The MLP-Mixer consists of a repeated unit called the mixer layer, which has two main components: the channel-mixing MLP and the token-mixing MLP. The channel-mixing MLP is similar to a depthwise convolution, designed for interaction between channels, while the token-mixing MLP facilitates interaction between spatial patches. Unlike transformer models, the MLP-Mixer does not require a complex self-attention mechanism and instead alternates between stacking the channel-mixing MLP and the token-mixing MLP. Despite their simpler structure, MLP-based models have achieved accuracy comparable to that of transformer models.


Inspired by recent advancements, several MLP-based models \cite{zhang2021morphmlp,yu2021rethinking} have rapidly emerged, showcasing tremendous potential. This motivates us to design an MLP-based model for ophthalmic image segmentation. However, we face two significant challenges in developing MLP-based ophthalmic segmentation models.
Firstly, unlike transformer architectures that incorporate a position embedding operator, MLP-based models can easily lose location information after several fully-connected layers. While this may not be problematic for classification tasks, maintaining location information is crucial for segmentation tasks.
Secondly, similar to transformer architectures, most MLP-based models require pre-training on large datasets to perform well, due to the absence of image-specific inductive biases. This presents a challenge, as ophthalmic datasets are typically small \cite{khan2021global}.


To address these challenges, we have designed a Multi-Scale MLP (MMLP) module and proposed a mixed MLP architecture termed MM-UNet. The hybrid design, integrating convolutional and MLP components, aims to leverage the strengths of both approaches. Initially, we utilize UNet \cite{ronneberger2015u} to extract multi-level local features. Subsequently, the MMLP module re-establishes long-range dependencies while partially retaining local information. Compared to the MLP-mixer, our MMLP module omits the channel-mixing MLP, as the convolutional components in our model already establish inter-channel relationships. Furthermore, the MMLP module groups channels into different scales for local token-mixing rather than employing global token-mixing, thereby constraining the token-mixing range to several defined sizes. Additionally, the MMLP module reduces computational consumption compared to the MLP-mixer. We evaluated MM-UNet using two datasets: one from the anterior segment and the other from the posterior segment. Experimental results demonstrate that our mixed MLP architecture exhibits significant potential in ophthalmic image segmentation.

\section{Method}
\subsection{Revisiting MLP-mixer Mechanism}
MLP-mixer is a pioneering architecture that shows the potential of the pure-MLP models. It consists of three parts: per-patch embedding layer, Mixer layers, and classification layer. In this section, we briefly review this inspiring pure-MLP method. 

\textbf{Per-patch embedding layer.} For an given image input $I\in\mathbb{R}^{3\times W\times H}$,  we firstly crop $I$ into $n$ non-overlap patches $p_{i} \in \mathbb{R}^{P\times P\times 3}$, where $n=WH/P^{2}$. After that, each $p_{i}$ is unfolded into a vector in the space of $\mathbb{R}^{3P^{2}}$. At this time, a shared fully connected layer named the embedding layer projects each patch $p_{i}$ into a hidden dimension $C$. Now we get an output $X\in \mathbb{R}^{n \times C}$ that represents $n$ patches with dimension $C$.

\textbf{Mixer layers.}
The mixer layer is built upon two types of MLP layers: the channel mixing MLP and the token mixing MLP. The former realizes the information exchange between channels, and the latter realizes the establishment of tokens. For the input $X$ calculated by the embedding layer, the mixer layers can be written as follows:

\begin{equation}
U=U+W_{2}\sigma[W_{1}LayerNorm(X)],
\label{eq2}
\end{equation}

\begin{equation}
Y=U+\sigma(LayerNorm(U)W_{3})W_{4}. 
\label{eq3}
\end{equation}

Here, the Eq. \ref{eq2} represents channel mixing, $W_{1}$ is the weights of a fully-connected layer increase the feature dimension by a ratio $r$. Moreover, $W_{2}$ denotes a subsequent fully-connected layer that reduces the feature dimension back to the original size. Moreover, $LayerNorm(\cdot)$ represent the layer normalization \cite{ba2016layer} and $\sigma(\cdot)$ denotes the nonlinear activate function GELU\cite{hendrycks2016gaussian}. Equation \ref{eq3} represents token mixing, which is similar to channel mixing, except changing the target dimension from channel to block.

\textbf{Clasification layer.}
After a repeat of $N$ mixer layers, we suppose the final output of mixer layers is $Y \in \mathbb{R}^{n \times d}$. The classification layer can be formulated as follow: 

\begin{equation}
Classifier(Y)=fc(GAP(Y)), \label{eq4}
\end{equation}

where $GAP(\cdot)$ represents global average pooling, and $fc(\cdot)$ denote the fully-connected layer.

\begin{figure}[t]
\includegraphics[width=0.95\textwidth]{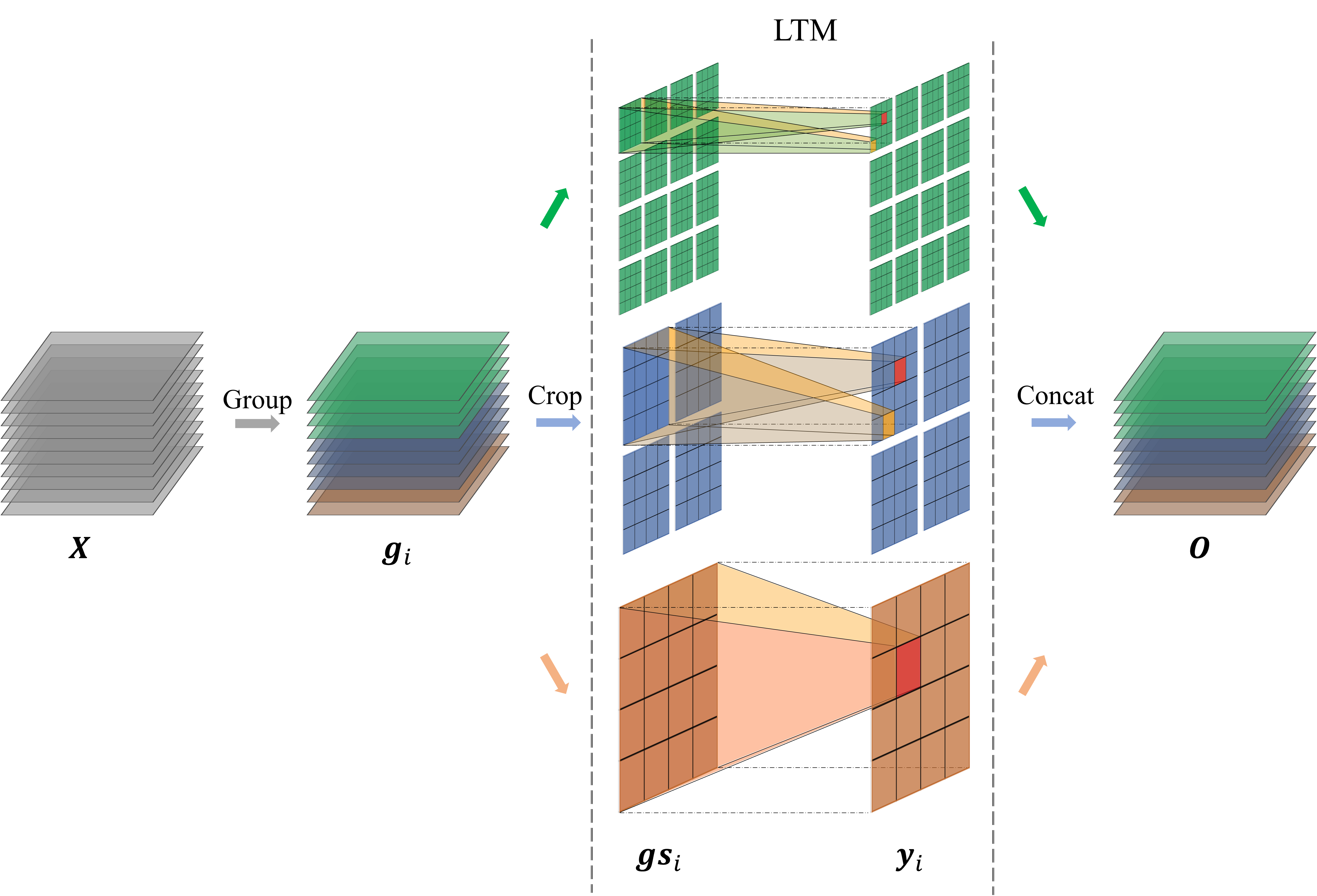}
\caption{The proposed multi-scale MLP (MMLP) block.} \label{MMLP}
\end{figure}

It is evident that the overall design of the MLP-mixer is remarkably simple, and its robust capability to model long-range dependencies primarily stems from the token-mixing operation.

\subsection{Multi-scale MLP Block}
Upon revisiting the concept, we recognize the potential of token-mixing MLP. However, the token-mixing step also completely intermixes spatial information, resulting in the loss of location information. While location information may not be crucial for classification tasks, it is highly significant for segmentation. To preserve location information during the token-mixing process, we have made improvements to the original token-mixing operator and proposed a new MLP structure, the Multi-Scale MLP (MMLP), as illustrated in Fig.~\ref{MMLP}.

For an input $X\in\mathbb{R}^{C\times W\times H}$, we first divide $X$ into k groups by channel, denoted as  

\begin{equation}
\begin{aligned}
& g_{1}=X[0:i_{1},:,:], \\
& g_{j}=X[i_{j-1}:i_{j},:,:], \\
& g_{k}=X[i_{k-1}:C+1,:,:], \\
& 0<i_{1}<i_{2}<\dots<i_{k-1}<C+1, \label{eq5}
\end{aligned}
\end{equation}

where C is the number of channels, and $g_{i},1\leq i \leq k$ denotes the $i-th$ group. For the $i-th$ group $g_{i}$, the mixing method is defined follow:

\begin{equation}
\begin{aligned}
& gs_{i} = crop(g_{i}, s_{i}) \\
& y_{i} = LTM(gs_{i}, n_{i}) 
\end{aligned}
\end{equation}

Here, $crop(\cdot)$ refers to cropping a feature map into several $s_{i} \times s_{i}$ patches, similar to the process in Mixer-MLP. Unlike the global correlations established by the token mixing mentioned above, $LTM(\cdot)$ is a local token-mixing operator which divides the entire space into $n_{i} \times n_{i}$ ($n_{i} \in [1, \frac{W}{s_{i}}]$) blocks and then establishes interactions within each block which is represented in Fig.~\ref{MMLP}. 

When $n_{i}=1$, it is equivalent to a traditional global token mixing. 

Otherwise, when $n_{i}\neq 1$, only local dependencies are established, resembling a convolution operator. We manually apply different parameters $\{s_{i}, n_{i}\}$ for different group of features $g_{i}$. Finally, we get the obtain the output $o$ using a concatenate operator $Concat(\cdot)$:

\begin{equation}
    o = Concat([y_{1}, y_{2}, y_{3},\dots, y_{k}])
\end{equation}

The MMLP block can capture multi-scale receptive fields via the sequence of local token-mixing operators. For some channels set with a large $n_{i}$ preserve more position information, the other channels with a small $n_{i}$ prefer to establish the long-range relationship.

\begin{figure}[t]
\includegraphics[width=0.95\textwidth]{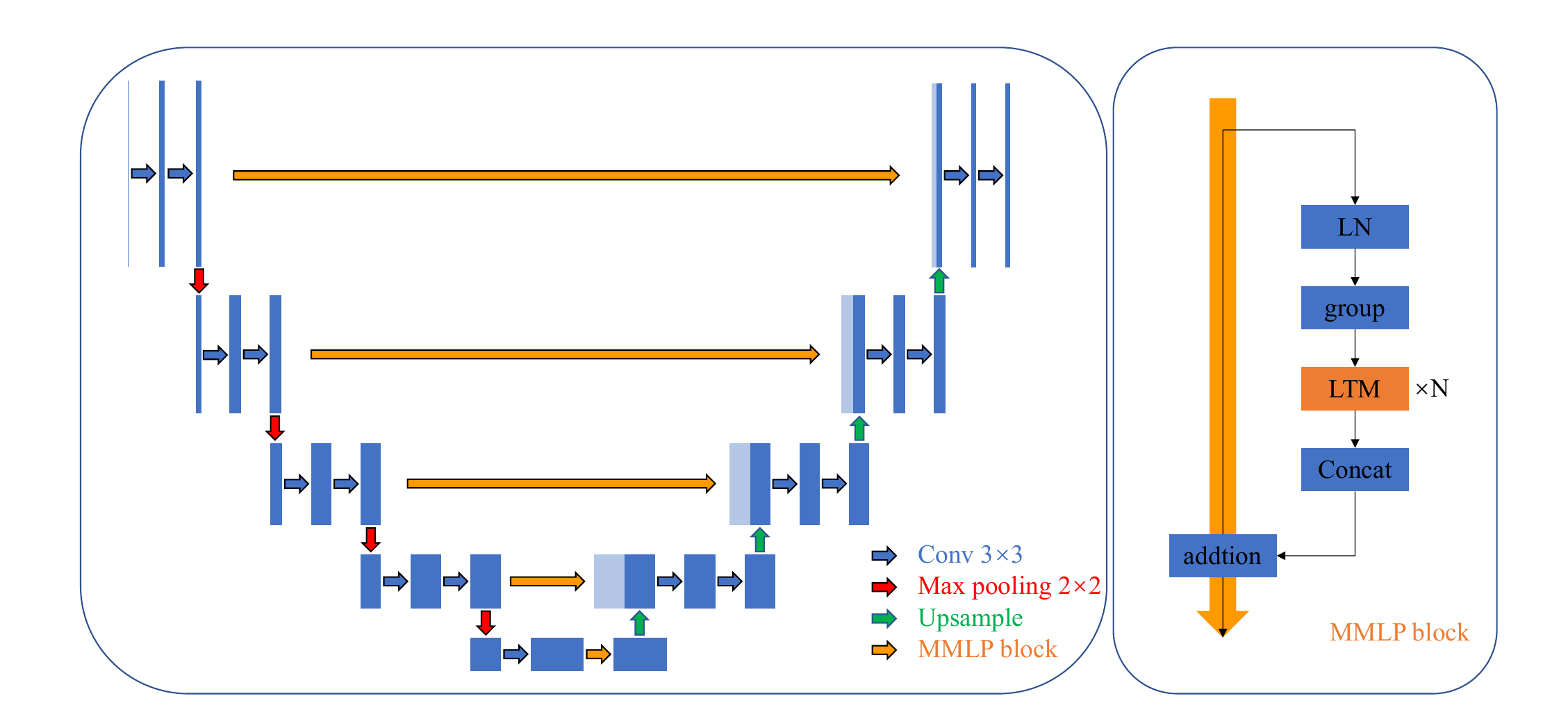}
\caption{The scheme of MM-Unet.} \label{MM-Unet}
\end{figure}

\begin{table}[b]
\caption{The parameter settings in MM-UNet.}\label{set}
\centering
\begin{tabular}{c|c|c|c}
\hline
Layer & Channel Group & $n_{i}$ & $s_{i}$ \\
\hline
1 & [32,16,16] & [32,16,8] & 4 \\
2 & [64,32,32] & [16,8,4] & 4 \\
3 & [128,64,64] & [8,4,2] & 4 \\
4 & [256,128,128] & [8,4,2] & 4 \\
5 & [512,256,256] & [4,2,1] & 4 \\

\hline
\end{tabular}
\end{table}

\subsection{Network Architecture}

We proposed a mixed MLP architecture that combines the MLP and UNet architectures, termed MM-UNet. Figure~\ref{MM-Unet} illustrates the architecture of MM-UNet. Our objective is to harness the strengths of both convolutional and MLP mechanisms. This hybrid architecture leverages the inductive bias of images inherent in convolutional operations to avoid the need for large-scale pre-training, while also utilizing the MLP's capability to capture long-range relationships. Since segmentation is essentially a pixel-level classification task, preserving positional information during processing is crucial. Therefore, we retain the original encoder and decoder to prevent the loss of location information. Given that images are highly structured data, low-level features contain more localized information. Consequently, we configure larger $n_{i}$ values in the MMLP block at lower levels compared to higher levels. The specific settings are detailed in Table~\ref{set}.

\section{Experiment}

\subsection{Dataset}

We selected two modalities of commonly used ophthalmic images, the Anterior Segment Optical Coherence Tomography (AS-OCT) and fundus, to verify the effectiveness of MM-UNet.

\begin{figure}[t]
\centering                                                          
\subfigure[]{                    
\begin{minipage}[t]{0.23\textwidth}
\centering                                                          
\includegraphics[scale=0.4]{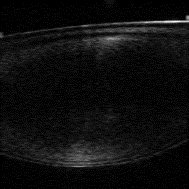}               
\end{minipage}}
\subfigure[]{                    
\begin{minipage}[t]{0.23\textwidth}
\centering
\includegraphics[scale=0.4]{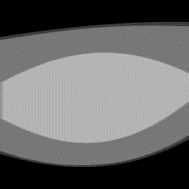}                
\end{minipage}}
\subfigure[]{                    
\begin{minipage}[t]{0.23\textwidth}
\centering
\includegraphics[scale=0.4]{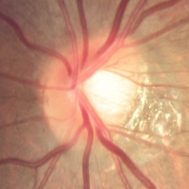}                
\end{minipage}}
\subfigure[]{                    
\begin{minipage}[t]{0.23\textwidth}
\centering
\includegraphics[scale=0.4]{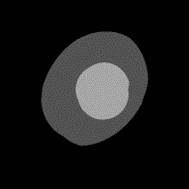}                
\end{minipage}}
\caption{This represents the two ophthalmic datasets we used. (a) and (b) is the AS-OCT image and its segmentation label, respectively; (c) and (d) show the fundus image of the REFUGE2 dataset.} 
\label{dataset}
\end{figure}

\textbf{AS-OCT Dataset.} The AS-OCT is our private dataset collected through the CASIA2 (Tomey Corporation, Japan) Ophthalmic Imaging device, with the label of lens substructure, which is shown in Fig. \ref{dataset}. Our task is to divide the lens area into nucleus, cortex, and capsule, which is crucial for the surgical judgment of cataracts. There are 1844 images here from 284 subjects, including 154 cataract lenses and 130 healthy lenses. We label four images with equal intervals for each eye. So we labeled 461 eyes, which contain 230 right eyes and 231 left eyes, and some images are missing due to the occlusion of the eyelids during collection.

\textbf{Fundus Dataset.} We use the REFUG2 fundus dataset in MICCAI 2019\cite{fang2022refuge2}, which is a multi-domain dataset collected from different devices. Although the entire REFUGE2 dataset contains 2000 images with multi-task labels for segmentation, classification, and localization, we only used the available 1200 fundus images for the optic cup and optic disc segmentation task. 

\subsection{Experiment Setup}
For data division, we divide the AS-OCT dataset by subject into three disjoint subsets: training dataset, validation dataset, and testing dataset in a ratio of 6:2:2 on a random selection basis. Furthermore, in the REFUGE2 dataset, we preserve the original division strategy: 400 images for training, 400 images for validation, and 400 images for testing.   
We train all models from scratch for 150 epochs. We use the stairs learning rate scheduler with a base
learning rate of 0.015, which decreased by 10 every 10 epochs after 100 epochs. We use the SGD optimizer with a batch size of 16 on one 12G Titan V GPU with a momentum of 0.9 and a weight decay of $10^{-4}$, respectively. All images are resized to 256 × 256 before entering the models. We use the cross-entropy loss to train all models.

We use the mIoU value to evaluate the segmentation performance of the model. The calculation formulas are as follows: 

\begin{equation}
    mIoU=\frac{1}{N}\sum_{i=1}^{N}\frac{|A_{i}\cap B_{i}|} {|A_{i}\cup B_{i}|},
\end{equation}

where $A_{i}$ represents the model prediction area for category $i$, $B_{i}$ represents the ground truth area for category $i$, and $N$ is the total number of categories. Furthermore, we use $\#P$ to denote the number of parameters and $Acc$ to represent accuracy. 

\subsection{Segmentation Results}

\begin{figure}[t]
\includegraphics[width=\textwidth]{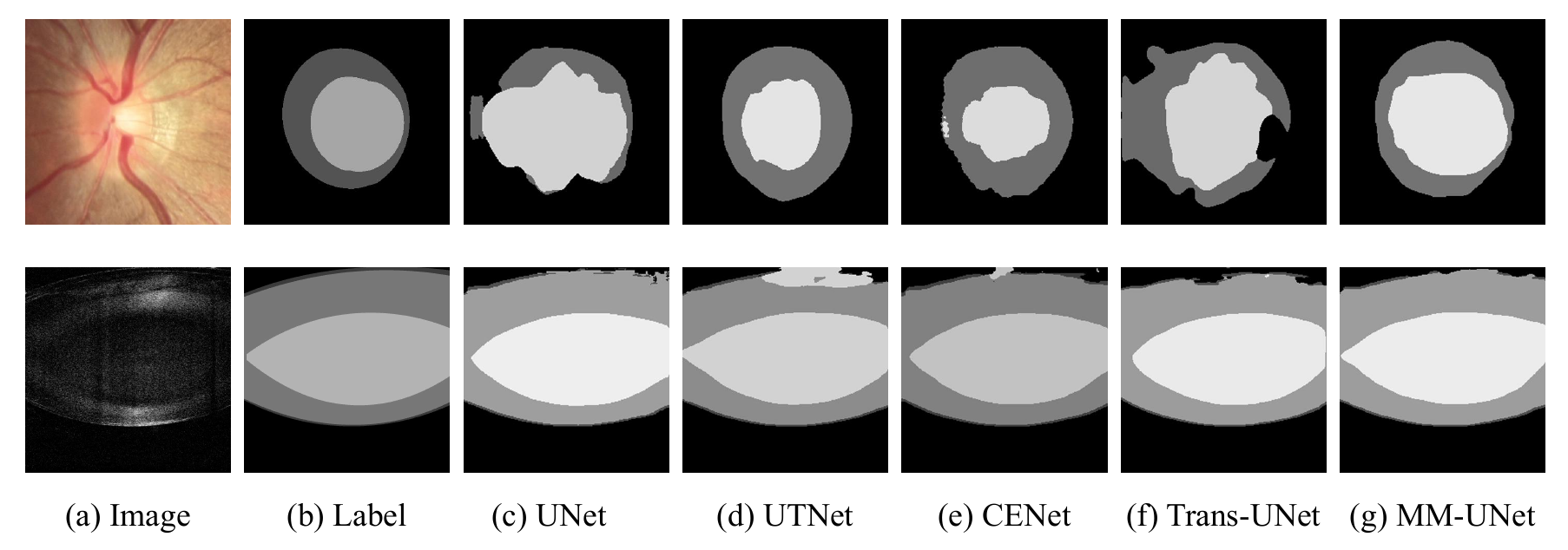}
\caption{The comparison results of our proposed MM-UNet with other state-of-the-art models.} \label{r}
\end{figure}

Table~\ref{tab1} shows the compared results of our proposed MM-UNet with other CNN-based and transformer-based models. We can see that our proposed MM-UNet achieves the best performance both on the AS-OCT dataset and on the REFUGE2 dataset. In the AS-OCT dataset, our method achieves $98.2\%$ accuracy and $92.64\%$ mIoU, respectively, which outperforms  $1.2\%$ mIoU than UNet with almost no parameter increase. In the REFUGE2 dataset, MM-UNet also improves performance with a low-level parameter. Fig.~\ref{r} shows some of the segmentation results, and we can easily find MM-Unet performers well in the uncertainty boundary compared with other methods.

\begin{table}
\caption{Comparison with state-of-the-art segmentation methods.}\label{tab1}
\centering
\begin{tabular}{c|c|cccccc}
\hline
Method &  Dataset & Acc(/\%) & mIoU(/\%) & \#P \\
 \hline
UNet\cite{ronneberger2015u} & \multirow{5}*{AS-OCT} & 97.91 & 91.43 &\textbf{12.77M}\\
CENet\cite{gu2019net} & & 98.07 & 91.69 & 27.66M \\
Trans-Unet\cite{chen2021transunet} & & 97.89 & 91.82 & 88.88M \\
UTNet\cite{gao2021utnet} & & 98.13 & 92.36 &15.55M\\
MM-UNet(ours) & & \textbf{98.20} & \textbf{92.64} &12.98M \\
 \hline
UNet\cite{ronneberger2015u} & \multirow{5}*{REFUGE2} & 91.24 & 74.36 &\textbf{12.77M} \\
CENet\cite{gu2019net} & & 93.58 & 79.77 & 27.66M\\
trans-Unet\cite{chen2021transunet} & & 93.17 & 78.48 & 88.88M\\
UTNet\cite{gao2021utnet} & & 93.34 & 78.70 &15.55M \\
MM-UNet(ours) & & \textbf{93.91} & \textbf{80.41} &12.98M \\
 
\hline
\end{tabular}
\end{table}

\subsection{Ablation Study}
To verify the effectiveness of our proposed MMLP block, we tried to replace the LTM operator with the global token-mixing MLP. Moreover, the compared result is shown in Tab.~\ref{tab2}. 
To verify the effectiveness of our proposed MMLP block, we replaced the LTM operator with the global token-mixing MLP. The comparative results are presented in Table~\ref{tab2}. Our LTM demonstrates improvements in both mIoU and Acc compared to the token-mixing methods in both AS-OCT and REFUGE2.

\begin{table}
\caption{The impact of different token-mixing operations on segmentation results.}\label{tab2}
\centering
\begin{tabular}{c|c|cccccc}
\hline
MLP operator &  Dataset & Acc(/\%) & mIoU(/\%) & \#P \\
 \hline
token-mixing & \multirow{2}*{AS-OCT} &  98.04 & 92.15 &19.17M\\
LTM && 98.20 & 92.64 &12.98M\\
\hline
token-mixing & \multirow{2}*{REFUGE2} &  93.66 & 79.00 &19.17M\\
LTM &  & 93.91 & 80.41 &12.98M\\ 
\hline
\end{tabular}
\end{table}

\section{Conclusion}

The growing popularity of MLP architecture underscores its significant potential in various MLP-based applications. In this study, we introduce a novel mixed MLP architecture specifically designed for ophthalmic image segmentation, representing a pioneering effort in the field of ophthalmology. Furthermore, we developed a Multi-Scale MLP (MMLP) module to enhance segmentation performance. Experiments on the private AS-OCT dataset and the REFUGE2 dataset demonstrate that our hybrid MLP architecture significantly improves the model's capacity to capture long-distance dependencies. Looking ahead, we intend to extend our MLP-based approach to a broader range of applications.

\section*{Acknowledgements}
This work was supported in part by National Natural Science Foundation of China (No.82272086).

\bibliographystyle{splncs04}
\bibliography{mybibliography}

\end{document}